\documentclass[5p,times,procedia]{elsarticle}
\usepackage{ecrc}
\usepackage{footnote}
\usepackage{verbatim}

\volume{00}

\firstpage{1}

\journalname{Procedia CIRP}

\runauth{Chengzhi Wu et al.}

\jid{trpro}

\usepackage{amssymb}

\usepackage{wrapfig}
\usepackage{subcaption}
\usepackage[bookmarks=false]{hyperref}
    \hypersetup{colorlinks,
      linkcolor=blue,
      citecolor=blue,
      urlcolor=blue}

\usepackage{ulem}

\begin{document}
\begin{frontmatter}

\dochead{9th CIRP Conference on Assembly Technology and Systems}%

\title{MotorFactory: A Blender Add-on for Large Dataset Generation \\ of Small Electric Motors}

\author[a]{Chengzhi Wu *}
\author[a]{Kanran Zhou}
\author[b]{Jan-Philipp Kaiser}
\author[c]{Norbert Mitschke}
\author[d]{Jan-Felix Klein}
\author[e]{\\Julius Pfrommer}
\author[a,e]{Jürgen Beyerer}
\author[b]{Gisela Lanza}
\author[c]{Michael Heizmann}
\author[d]{Kai Furmans}

\address[a]{Institute for Anthropomatics and Robotics, Karlsruhe Institute of Technology, Kaiserstraße 12, 76131 Karlsruhe, Germany}
\address[b]{wbk Institute of Production Science, Karlsruhe Institute of Technology, Kaiserstraße 12, 76131 Karlsruhe, Germany}
\address[c]{Institute of Industrial Information Technology, Karlsruhe Institute of Technology, Hertzstraße 16, 76187 Karlsruhe, Germany}
\address[d]{Institute for Material Handling and Logistics, Karlsruhe Institute of Technology, Kaiserstraße 12, 76131 Karlsruhe, Germany}
\address[e]{Fraunhofer Institute of Optronics, System Technologies and Image Exploitation IOSB, Fraunhoferstraße 1, 76131 Karlsruhe, Germany}

\aucores{* Corresponding author. Tel.: +49-(0)1523 8476995. {\it E-mail address:} chengzhi.wu@kit.edu}

\begin{abstract}
To enable automatic disassembly of different product types with uncertain condition and degree of wear in remanufacturing, agile production systems that can adapt dynamically to changing requirements are needed. Machine learning algorithms can be employed due to their generalization capabilities of learning from various types and variants of products. However, in reality, datasets with a diversity of samples that can be used to train models are difficult to obtain in the initial period. This may cause bad performances when the system tries to adapt to new unseen input data in the future. In order to generate large datasets for different learning purposes, in our project, we present a Blender add-on named MotorFactory to generate customized mesh models of various motor instances. MotorFactory allows to create mesh models which, complemented with additional add-ons, can be further used to create synthetic RGB images, depth images, normal images, segmentation ground truth masks and 3D point cloud datasets with point-wise semantic labels. The created synthetic datasets may be used for various tasks including motor type classification, object detection for decentralized material transfer tasks, part segmentation for disassembly and handling tasks, or even reinforcement learning-based robotics control or view-planning.
\end{abstract}


\end{frontmatter}

\section{Introduction}
\label{introduction}
Today's industrial landscape is characterised by linear economies. End-of-life (EOL) strategies for products such as remanufacturing, in which used products are reprocessed, offer the potential to decouple resource consumption from sustainable economic growth \cite{Tol17}. During the remanufacturing process, the products are inspected and disassembled, then individual parts are reworked or exchanged and again reassembled into final products \cite{Sun04}. In contrast to related EOL-strategies, remanufactured products ensure functionality and quality that is equivalent or better compared to a new product \cite{Tol17}. Remanufacturing systems face various challenges originating from uncertain product states, inconsistent quality and fluctuating availability of products. Consequently, even today the vast majority of processes in a remanufacturing system are carried out manually \cite{Kur18}. In order to automate these processes, agile production systems consisting of autonomously operating subsystem are required which provide the highest possible flexibility and adaptability. 

\begin{figure*}[t]
    \centering
    \includegraphics[width=0.9\textwidth]{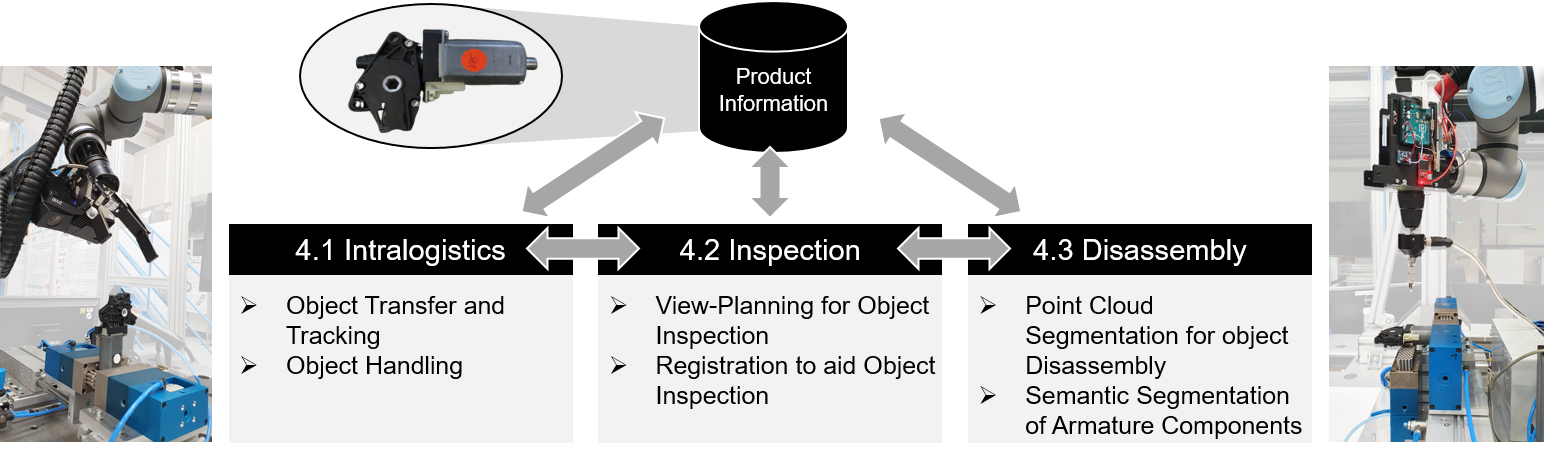}
    \caption{Product-based relationships and considered applications between intralogistics, inspection and disassembly in an automated remanufacturing system.}
    \label{fig:overview}
\end{figure*}

In this paper, we consider the automated disassembly of different variants of end-of-life actuators which are commonly used in vehicle manufacturing, e.g., as seat adjuster motors, window lift motors or rear door motors. As shown in figure \ref{fig:overview}, the considered subsystems are the intralogistics, the inspection, as well as the disassembly which all heavily rely on product information. However, in remanufacturing, each core is unique since products have a high variance concerning their product state. Therefore, handling, inspection and disassembly strategies can not be defined in advance. Accordingly, the system must derive and execute these strategies during runtime based on the actuator at hand. 
 
Machine learning methods, and especially deep learning methods, may be the key to achieve the necessary robustness to deal with the high degree of variability. By learning the internal structure on part level, (e.g. gear container, pole pot, electrical connection), processes on unseen variants which have similarities to the known population of actuators become feasible. However, a major disadvantage of machine learning methods is the required amount of data. Allocating and annotating a large amount of data is time-consuming or even impossible in reality. In the recent past, the synthetic generation of training data for machine learning methods has become a popular alternative. By leveraging transfer learning, it allows to train models that rely less on elaborately labeled real-world data. We therefore present an approach for generating synthetic training dataset as well as its relevant applications for the handling-, inspection-, and disassembly processes.

The remainder of this paper is structured as follows: Section \ref{sec:stateoftheart} summarizes the state of the art of 3D synthetic dataset creation and application. Section \ref{sec:add-on} describes technical details on the developed Blender add-on. Section \ref{sec:applications} provides brief overviews on applications build on top of the add-on while section \ref{sec:conclusion} summarizes presented outcome and discusses future work.

\section{State of the art}
\label{sec:stateoftheart}
Since the appearance of the iconic \textit{Stanford Bunny}, generating synthetic datasets as training data for machine learning purposes has already been widely discussed and used as a possible learning approach for various computer vision applications. 
Regarding synthetic dataset of 3D models, the Princeton Shape Benchmark \cite{Shilane2004ThePS} provides a collection of 1,814 polygonal models of objects from different categories as an early dataset. ModelNet \cite{Wu20153DSA} contains 127,915 models for 3D object classification and retrieval. More than three million annotated 3D models are collected in ShapeNet \cite{Chang2015ShapeNetAI}. Its subsequent work PartNet \cite{mo2018partnet} additionally offers fine-grained semantic segmentation information for a subset of the models. By utilizing Thingiverse, Thingi10K \cite{Zhou2016Thingi10KAD} provides a large dataset of 3D-printing models. More recently, the ABC dataset \cite{Koch2019ABCAB} 
collects over 1 million CAD models, including lots of  mechanical parts with sharp edges and well defined surfaces, which are seldom included in the previous synthetic datasets.

Regarding 3D scenes, \cite{Var17} generates a synthetic dataset for the segmentation and detection of objects in virtual street scenes. Also \cite{Ros16} and \cite{Kha19} consider urban scenes and each provides a dataset for semantic segmentation in these environments. Other approaches in the image domain deal with the generation of images from garden scenes \cite{Hoa20}, or specifically for object detection and pose estimation \cite{Trem18}. There are also approaches for the generation of point clouds, such as that of \cite{Gri19}, in which, in contrast to the previously mentioned work, point clouds of urban scenes are generated using Blender. Another work using Blender deals with automatic generation of point clouds of historical objects \cite{Pie19}. 

However, in the production environment for industrial applications, approaches of generating synthetic dataset are seldom employed. They may contribute in various applications including product classification, segmentation of product components, product tracking, and even determination of grasp points.

\section{MotorFactory Add-on}
\label{sec:add-on}

\begin{figure}[t]
    \centering
    \includegraphics[width=0.85\columnwidth]{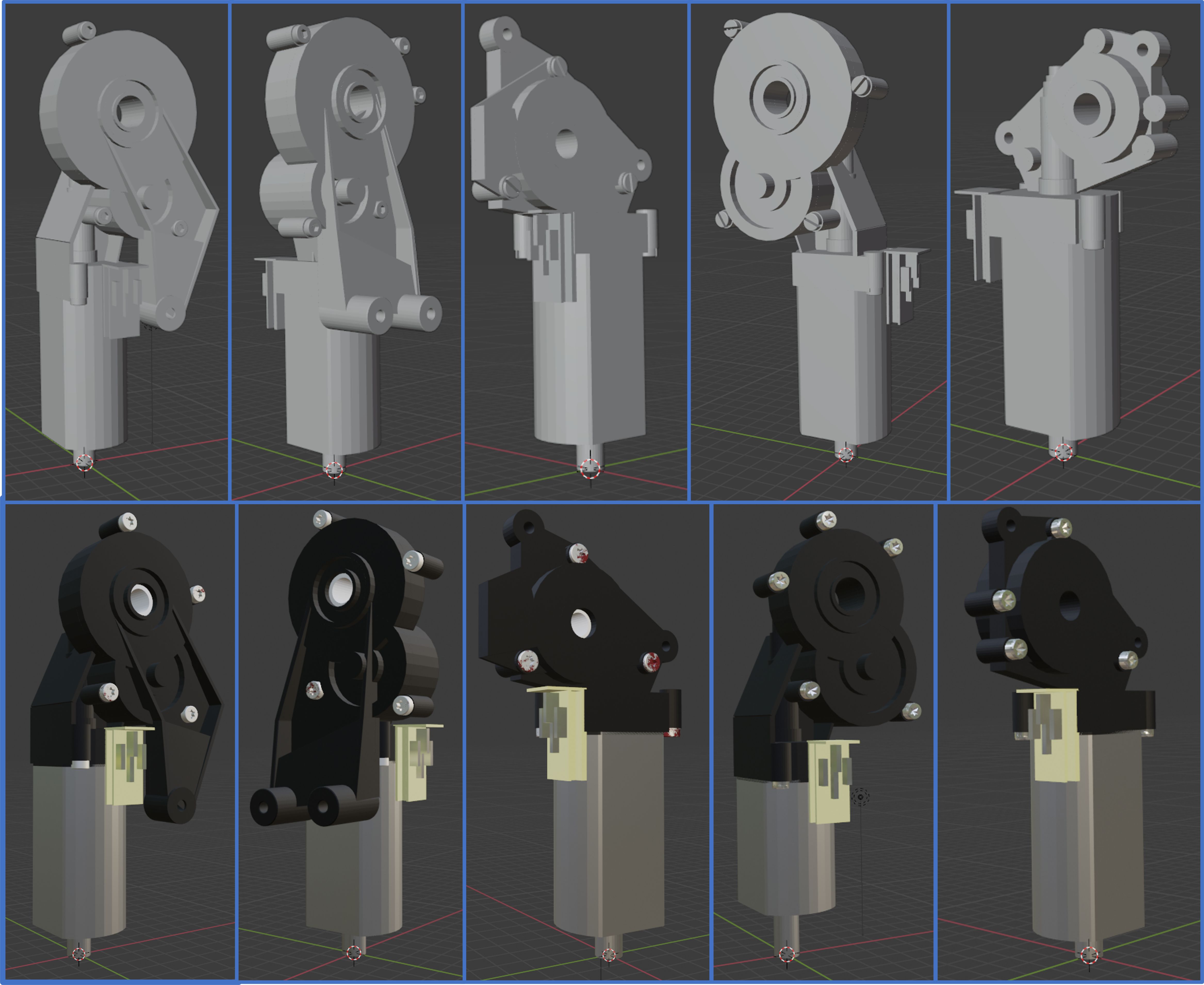}
    \caption{Generated demo motors. Upper row: no textures added; bottom row: textures added and rendered. Column 1/2/4: Type-A motors with two gears; column 3/5: Type-B motors with one gear.}
    \label{fig:demo1}
\end{figure}

\begin{figure}[t]
    \centering
    \includegraphics[width=\columnwidth]{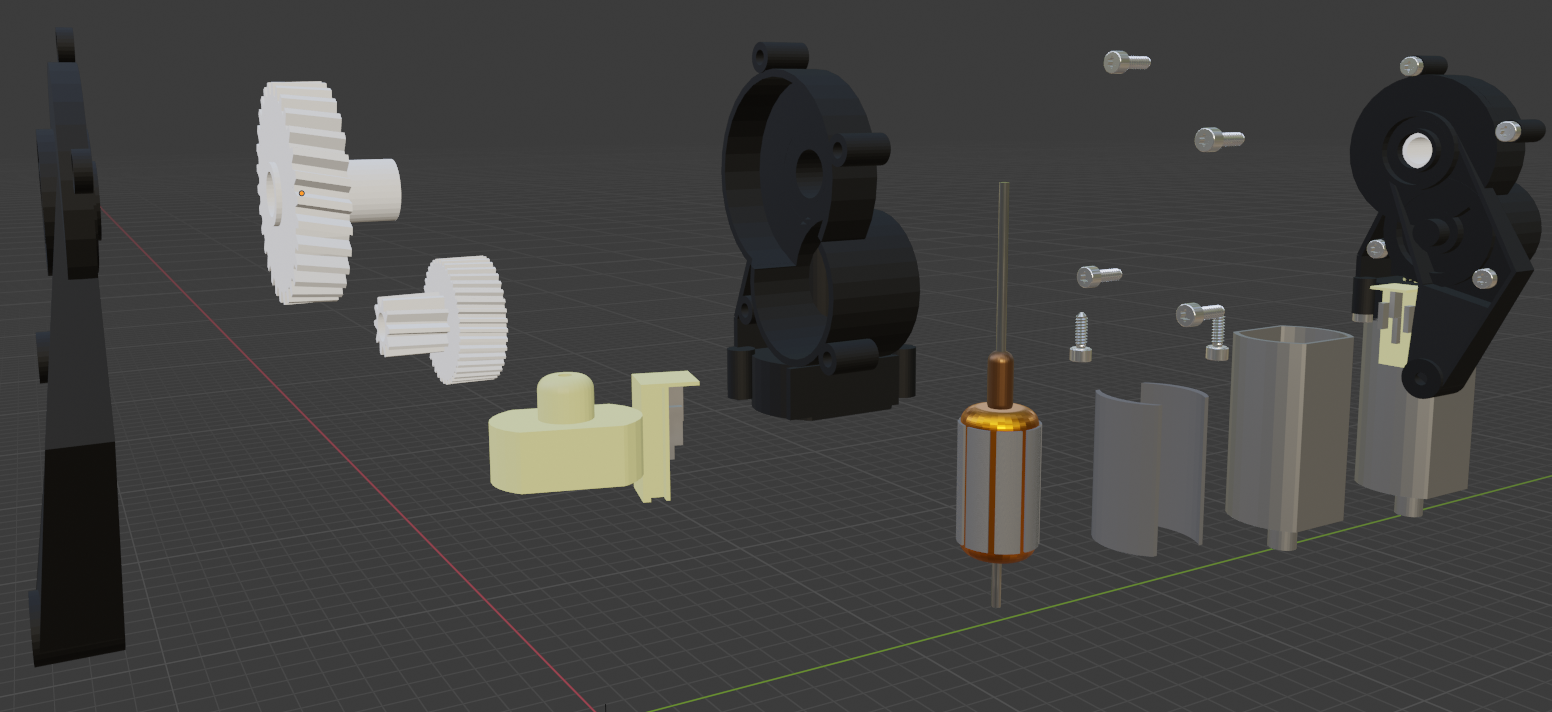}
    \caption{An explosion figure of a motor generated with Version 2.0. The original assembled motor model is also shown at the right most.}
    \label{fig:demo2}
\end{figure}

The AgiProbot project aims for auto-detection, tracking and disassembly of end-of-life products. To be specific, in our current experimental setting, we work on small electric motors used in vehicle manufacturing. Universal method needs to be developed to deal with various types of motors, including the ones with unseen specifications. However, we are only provided with a handful of motors with only few different product specifications. The variance of data is actually not sufficient for training with machine learning methods, especially those deep learning-based ones. To deal with this problem, we created a Blender add-on named MotorFactory, which can generate motor mesh models with a variety of specifications based on the motor types we have currently. 

As an open source software, Blender is a proven tool that performs well in modeling shapes and creating highly customizable add-ons. 
Our MotorFactory add-on is able to generate mesh models with various specifications and save them in desired file formats. Each component of a generated motor can also be saved separately. Considering different requirements, the add-on is implemented in two versions. Version 1.0 only considers the components of motors that can be directly observed from the appearance, while Version 2.0 further considers the inner components. The generated models of both versions contain the following components: (i) Pole Pot; (ii) Electric Connection; (iii) Gear Container; (iv) Cover and (v) Bolts. Regarding inner components in version 2.0, the additionally generated parts are: (vi) Magnets; (vii) Armature; (viii) Lower Gear and (ix) Upper Gear. To generate motors with various specifications, we provide lots of parameter options that control the type, size, position and rotation of different parts of motor, e.g. bolt position, gear size, or pole pot length. Additionally, both versions provide multiple bolt generation options to meet different requirements. 

To better design the building graph of different types of motors, we define two basic types based on the number of gears in a motor. Type-A motor indicates the motors with two gears inside, while Type-B motor indicates the motors with only one gear. Each motor type has different kind of gear containers. Different gear containers further have different covers, which come with different mounting points. Three options of extension shapes for covers are provided for Type-A motors, while two options of are provided for Type-B motors. Figure \ref{fig:demo1} shows ten generated demo motors with different parameters. Figure \ref{fig:demo2} shows an exploded view of a demo motor generated with Version 2.0. All the individual components mentioned above are modeled separately as illustrated.

\begin{figure}[t]
\centering
  \begin{subfigure}[b]{0.9\columnwidth}
    \centering
    \includegraphics[width=\linewidth]{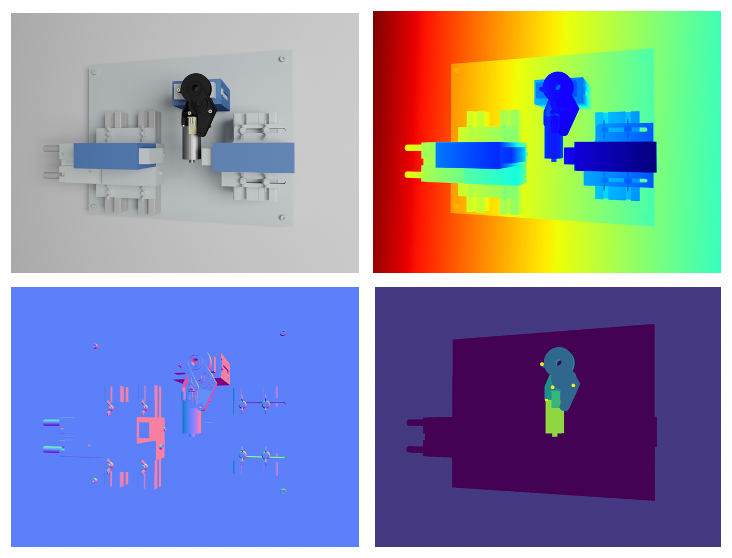}
    \caption{Demo of generated image dataset}
    \label{fig:imageDemo}
  \end{subfigure}
  \hfill 
  \begin{subfigure}[b]{\columnwidth}
    \centering
    \includegraphics[width=\linewidth]{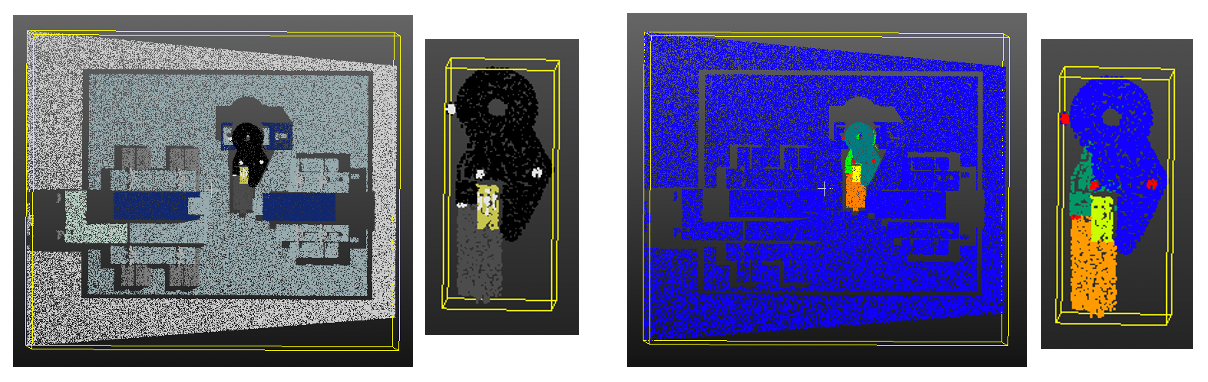}
    \caption{Demo of generated point cloud dataset}
    \label{fig:pcDemo}
  \end{subfigure}
\caption{Demos of generated dataset: (a) a rendered scene image with its corresponding depth image, normal image, and segmentation ground truth image; (b) a simulated scene point cloud with its point-wise segmentation ground truth.}
\label{fig:datasetDemo}
\end{figure}

The MotorFactory can be used to create a large amount of different motor variants which are further used to generate synthetic image and point cloud datasets. For example, to create an image datasets, apart from the scene images rendered by Blender directly, we can use BlenderProc \cite{denninger2019blenderproc} to generate the corresponding depth images, normal images, and segmentation ground truth images as illustrated in figure \ref{fig:imageDemo}. To create point cloud datasets, we can use Blensor \cite{Gschwandtner2011BlenSorBS} to simulate the sensors. Figure \ref{fig:pcDemo} gives a demo of generated scene point cloud with its segmentation ground truth. The dataset generation setting and pipeline depend on the actual needs of different tasks, hence they may vary from task to task.

\section{Application Use-Cases for the Motor Factory Add-On}
\label{sec:applications}
The ability to generate synthetic motor data for computer vision tasks enables the development or improvement of a variety of different use-cases inside the interdisciplinary AgiProbot project. The presented use-cases follow an exemplary product flow inside the AgiProbot demo-factory as illustrated in figure \ref{fig:overview}. The intralogistics system is responsible to transfer the product to a station and handle it with a vision based manipulator (\ref{sec:handling}). Depending on the station type, inspection related tasks (\ref{sec:view-planning}) or disassembly procedures (\ref{sec:pc_segmentation})  are carried out.

\subsection{Intralogistics}
\begin{figure}[t]
\centering
  \begin{subfigure}[b]{0.45\columnwidth}
    \centering
    \includegraphics[width=\linewidth]{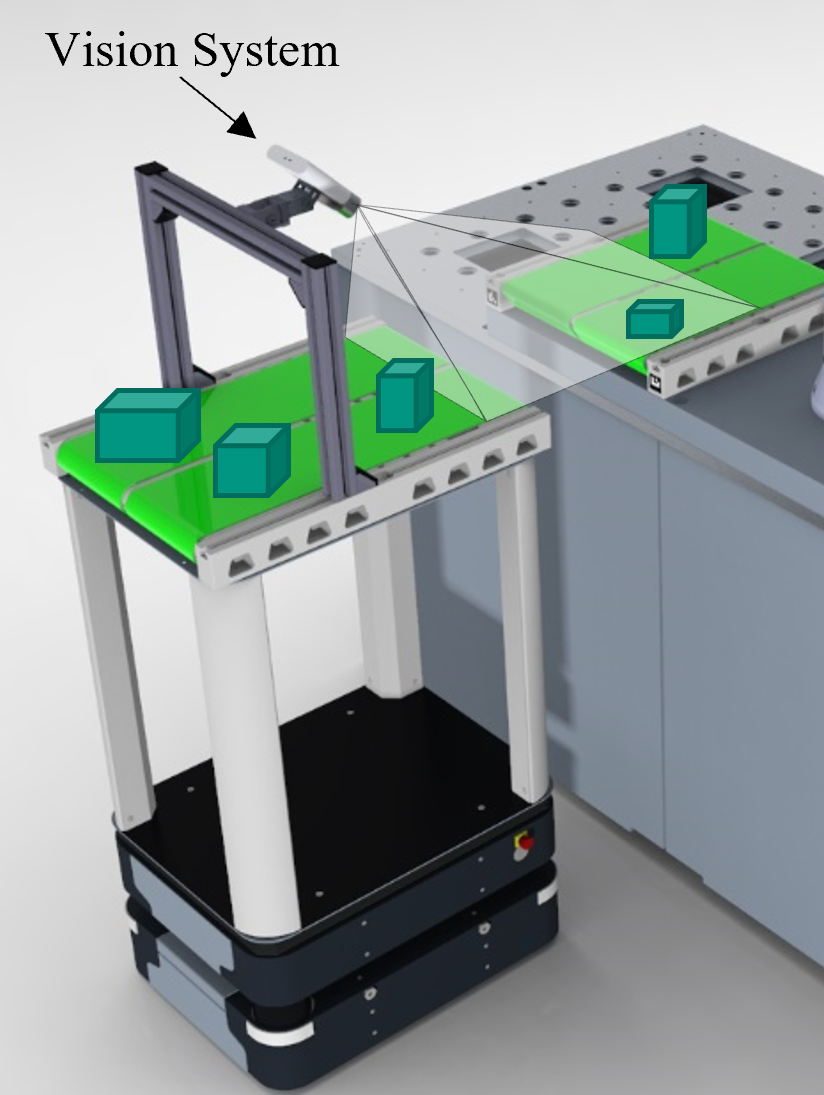}
    \caption{Object transfer scenario}
    \label{fig:material_transfer}
  \end{subfigure}
  \hspace{0.5cm} 
  \begin{subfigure}[b]{0.45\columnwidth}
    \centering
    \includegraphics[width=\linewidth]{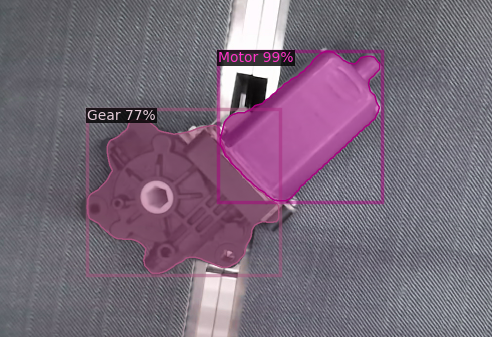}
    \caption{Instance segmentation}
    \label{fig:instance_segmentation}
  \end{subfigure}
\caption{Object transfer and handling, using a motor of Type-A as an example.}
\label{fig:fig}
\end{figure}

\textbf{Object Transfer and Tracking: }
\label{sec:transfer}
The AgiProbot production system follows the Fluid Automation framework to achieve the required changeability needed for automated remanufacturing \cite{Wurster.2021}. Realizing the material flow between individual stations in a matrix layout is the main task of the embedded intralogistics system \cite{Klein.2021}. This includes the transfer of objects between an autonomous mobile robot (AMR) and a transfer unit mounted on a station module, as illustrated in figure \ref{fig:material_transfer}. From a material flow perspective, objects are components such as products, parts or assemblies or carrier devices such as boxes or workpiece holders. The object transfer control workflow includes a vision system mounted on the AMR which is used to detect and track objects on the transfer units. Detecting objects and estimating their pose is essential to realize the decentralized control between the AMR and the transfer unit. For known objects, 2D object detection algorithms can be used to estimate the object's bounding boxes and track them over time. However, well performing deep learning based approaches, e.g. YOLO \cite{Bochkovskiy2020YOLOv4OS} and its tracker DeepSort \cite{Wojke.2017}, require a vast amount of labeled input data for the training process. In our case, by feeding generated motor models to Blender and using a BlenderProc-based pipeline to generate synthetic image data, one can create high quality rendered images with accurate ground truth annotations \cite{denninger2019blenderproc}. Enriched with labeled real-world images, one can create a dataset of reasonable size to train well performing real-time detection and tracking network models.  

\textbf{Object Handling: }
\label{sec:handling}
Object handling describes vision-based robotic pick-and place tasks, which are commonly required to meet necessary preconditions to perform value-adding tasks like manufacturing, assembly or disassembly.
A robotic grasping system commonly consists of a robotic manipulator, an RGB-D camera, a gripper and a target object. The robotic grasping problem can be generally divided into three sub-problems, namely the grasp detection, the trajectory planning and the execution \cite{Kumra.2017}. 

Hereby, the grasp detection system is the key entry point and includes the target object localization, the object pose estimation, and the grasp estimation \cite{Du.2021}. To solve the localization and pose estimation problem for known objects, simple bounding-boxes, as described in the previous use-case, are not sophisticated enough. However, since the MotorFactory generates mesh models on part level, annotated instance segmentation masks can be generated. Training a segmentation network like Mask R-CNN \cite{He.2017} allows to detect pixel-level segmentation instances of classes. Being able to detect parts individually allows to handle sub-assemblies and deriving pre-defined grasping points on part level. Figure \ref{fig:instance_segmentation} shows the segmentation results on a real-world image with a network model solely trained on synthetic data. Trained an instance segmentation network on many artificially generated product variants increases its robustness in detecting individual parts when a new variant is exposed to the system. In an adaptive vision system, compared to objects that are categorized to be unknown which only rely on salient detection, a never seen motor variant can therefore be detected with the help of more sophisticated methods .

\subsection{Inspection}
\textbf{View-Planning for Object Inspection: }
\label{sec:view-planning}
Since wear and tear or defects can occur anywhere on a returning product, it is usually necessary to inspect the entire surface of the product. The formalized problem of completing product surface coverage using a robot-based optical system can be expressed as a view-planning problem. In the literature so far, there are model-based and non-model-based solution approaches \cite{Sco03}. Model-based approaches determine robot poses for sensor sensing  by pre-planning based on a prior model. Non-model-based approaches determine the next best robot pose for acquisition at runtime through maximizing certain criteria. However, in many cases, independent remanufacturers do not have product information such as mesh models of the product at hand due to confidentiality issues \cite{Has17}. This hinders efficient pre-planning of visual strategies for inspection.

With the help of this add-on, an important research gap can be approached. In remanufacturing, the product variants considered are mostly different in detail but similar in principle. By using the generated dataset and machine learning methods such as reinforcement learning, inspection strategies can be learned in the simulation and transferred to previously unknown but similar product variants under sufficient generalization capability. In this case, the reinforcement learning agent generates an action (pose of the robot's end effector in space) based on a state (currently recorded point cloud of the product) and receives a reward (e.g. based on the relative information gain in the form of new detected surfaces). The reinforcement learning agent is trained in the simulation until its performance is satisfactory. In theory, the amount of data required for validation in reality then becomes smaller and the inspection agent can be used to fulfil inspection goals defined by the reward signal for a large number of different but similar variants of electric motors.

\begin{figure}[t]
\centering
  \begin{subfigure}[b]{0.48\columnwidth}
    \centering
    \includegraphics[width=\linewidth]{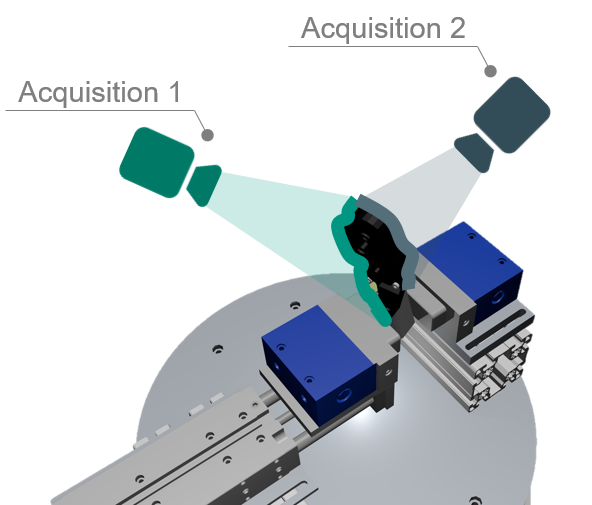}
    \caption{First clamping setup with the gear container facing upwards}
    \label{fig:VPP_1}
  \end{subfigure}
  \hfill 
  \begin{subfigure}[b]{0.48\columnwidth}
    \centering
    \includegraphics[width=\linewidth]{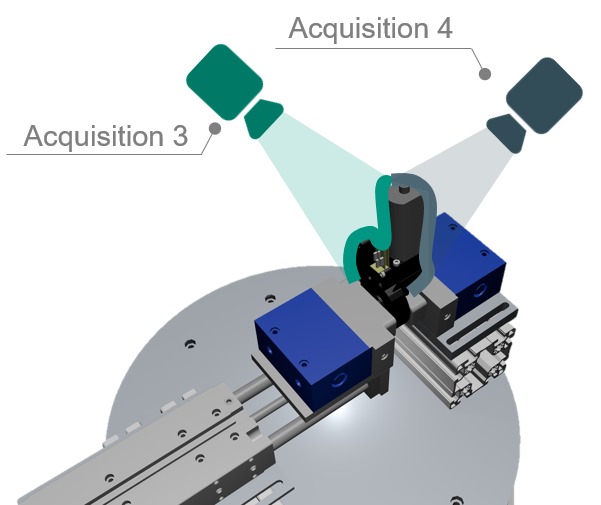}
    \caption{Second clamping setup with the pole pot facing upwards}
    \label{fig:VPP_2}
  \end{subfigure}
  \caption{Visualization of an exemplary inspection routine with two clamping setups to enable full target object inspection.}
  \label{fig:VPP}
\end{figure}

\textbf{Registration to aid Object Inspection: }
\label{sec:registration}
The requirements of completing surface coverage for objects can usually not be satisfied with one-time application of view-planning approaches. This is due to the fact that during inspection some surface areas of a product are permanently occluded by gripping or clamping units, or simply because the measured object is placed on a surface. After a successful product acquisition, for example with the approach described in view planning, a repositioning as well as the acquisition of the previously hidden surfaces is necessary. Since the repositioning changes the position of the object, a registration is necessary to merge the point clouds of all capturing steps in order to determine the entire object surface captured so far as well as to plan the next view based on. 

Random sample consensus (RANSAC)-based approaches are widely used for solving the registration problem, but they are model-free, computationally expensive and do not take the specific product properties into account \cite{Zho16}.  Here, the presented add-on offers the possibility to apply data-driven registration approaches with deep learning-based methods, which have proven successful compared to classical approaches \cite{Cho20}. Synthetic point clouds of a product model can be generated from different views of a sensor, whose pose is determined in the form of the position and the orientation. This also allows the transformation matrix between the acquired point clouds to be determined and can be used as a target variable for registration procedures to learn. In this way, product-specific properties can be taken into account during the training of the virtual registration process in order to achieve a faster and more precise registration result for real-world data.

\subsection{Disassembly}
\textbf{Point Cloud Segmentation for Object Disassembly:}
\label{sec:pc_segmentation}
Compared to images, point clouds contain more 3D information and is another widely used data representation in industrial applications. Point cloud segmentation provides an alternative way to segment the different components of a product to provide important information to robots. In recent years, methods of applying deep learning techniques on 3D point clouds have been intensively investigated. PointNet \cite{Qi2017PointNetDL} pioneered on this topic by designing a symmetric function for unordered point input, followed by works like PointNet++ \cite{Qi2017PointNetDH} or DGCNN \cite{Wang2019DynamicGC} which focused on gathering local information. Apart from classification, those methods also focus on point cloud segmentation including part segmentation on 3D shapes and scene segmentation for indoor scenes. 

\begin{figure}[t]
\centering
  \begin{subfigure}[b]{0.42\columnwidth}
    \centering
    \includegraphics[width=\linewidth]{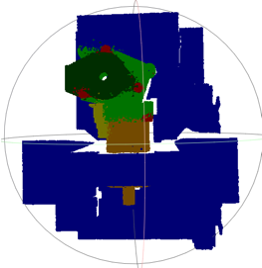}
    \caption{Synthetic point cloud}
    \label{fig:pcSeg_syn}
  \end{subfigure}
  \hspace{0.5cm} 
  \begin{subfigure}[b]{0.45\columnwidth}
    \centering
    \includegraphics[width=\linewidth]{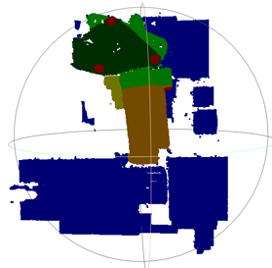}
    \caption{Real-world point cloud}
    \label{fig:pcSeg_real}
  \end{subfigure}
\caption{Comparison between (a) a synthetic point cloud with labels generated automatically, and (b) a real-world point cloud from the Zivid camera with points annotated manually. Using a motor of Type-B as an example.}
\label{fig:pcSeg}
\end{figure}

In our case, the target object is set fixed in a fixture when it is transferred to the disassembly station. Then a real-world point cloud is obtained with Zivid 3D Camera. However, the variance of different taken point clouds, especially regarding the motors, is not sufficient for learning purposes. In this case, we create a synthetic dataset using the proposed add-on. A demo is given in Figure \ref{fig:pcDemo}. Apart from the station table, the scene contains a fixure with a random motor placed in a relatively fixed position. To solve the problem of batches are not representative enough due to the large size of raw point clouds, we cut a cuboid of sub-point cloud that only considers the target object and its neighbor points after generating them. Further data augmentation is needed to enhance the generalization ability of the network, e.g. adding random small mask panels, adding random jitter noise, shifting fixure components, or rotating virtual cameras. Figure \ref{fig:pcSeg} gives a comparison between a synthetic point cloud and a real-world point cloud. Their scene settings are slightly different, but the synthetic one is already good enough to use as training data. After the network is pre-trained with the synthetic dataset, transfer learning may be applied with a few real-world data to finetune the model. Further important specification information may also be obtained with additional post-processing and subsequently provided to robots for other tasks.

\textbf{Semantic Segmentation of Armature Components:}
\label{sec:rotor_image_segmentation}
 The armature represents a functionally relevant part of the motor that requires a separate inspection in the given remanufacturing task. Semantic segmentation is required for the recognition of important surface parts of a armature. In a first proof-of-work approach, a small dataset of armatures has already been created to train a U-Net \cite{mitschke20b} for the segmentation task. However, the dataset contains only a small variety of armatures with 75 training images, most of them are taken from a similar perspective and are even partially inaccurately or inconsistently annotated.  Hence, the proposed Blender add-on can be used to generate additional training data to supplement or replace the given dataset. Different armature types and wear conditions can be obtained through suitable parameterization. Arbitrary perspectives on the armature can also be generated. Furthermore, the accurate annotation of the synthetic dataset improves the segmentation performance and also increases the meaningfulness of validation metrics.

\section{Conclusion and Outlook}
\label{sec:conclusion}
In this paper, we present MotorFactory, a blender add-on which allows the end-user to automatically create a variety of different small electric motor models and components. Based on the generated motors, synthetic datasets can be created for different computer vision tasks. Being able to generate and annotate product models easily may further increase the level of automation in challenging remanufacturing environments. Inside the interdisciplinary AgiProbot project, we identified several use-cases from the intralogistics, inspection, and disassembly  domains. All of the use-cases benefit greatly from having access to synthetic product models. 

In the future, an open-source benchmark dataset for object attribute estimation based on the add-on will be created and published. New versions of the presented add-on will include a greater amount of other electric motor types and variants, e.g. starters, or even other common remanufacturing products. Textures, which increase the level of realism, e.g. worn-out screw heads, rust or discolouration, are also intended to be added. The elaborated use-cases will be fully implemented and evaluated. Feedback based on the evaluation can then be used to further improve the add-on and its features.

\section*{Acknowledgements}
The AgiProbot project is funded by the Carl-Zeiss Foundation.

\bibliographystyle{style_bib} 
\bibliography{bibliography}

\clearpage\onecolumn

\end{document}